\documentclass{article}%[12pt]

\pdfoutput=1

\usepackage{multirow}
\usepackage{threeparttable}
\usepackage{makecell}
\usepackage{booktabs} 
\usepackage{amsmath,amsfonts,bm}

\def\eqref#1{Equation~(\ref{#1})}
\def\1{\bm{1}}

\RequirePackage{amsmath}
\RequirePackage{amssymb}
\RequirePackage{amsthm}
\RequirePackage{bm} 
\RequirePackage{url}
\usepackage{multirow}
\usepackage{natbib}
\usepackage{graphicx}
\usepackage{subfigure}
\usepackage{makecell}
\usepackage{booktabs}
\usepackage{array}
\usepackage{url}
\usepackage{algorithm}
\usepackage{algorithmic}
\usepackage{dsfont}

\usepackage{enumerate}
\usepackage[OT1]{fontenc}
\usepackage{natbib}

\usepackage{mathrsfs}

\usepackage{xcolor}
\usepackage{hyperref}
\usepackage{float}  % 放在导言区
\usepackage{colortbl}
\def\##1\#{\begin{align}#1\end{align}}
\def\$#1\${\begin{align*}#1\end{align*}}

\newcommand{\cD}{\mathcal{D}}

\newcommand{\E}{\mathbb{E}}

\usepackage{xcolor}

\definecolor{red1}{HTML}{f47983}
\definecolor{blue1}{HTML}{3eede7}
\definecolor{yellow1}{HTML}{f5dd6f}
\newcommand{\argmax}{\mathop{\mathrm{argmax}}}

\usepackage{hyperref}
\usepackage{array}
\usepackage{fullpage}
\usepackage{adjustbox}
\usepackage{lineno}

\title{
A Minimalist Approach to LLM Reasoning: from Rejection Sampling to Reinforce
}
\author{Wei Xiong$^*$$^\dagger$\quad Jiarui Yao$^\dagger$ \quad
Yuhui Xu$^\ddagger$\quad Bo Pang$^\ddagger$\quad  Lei Wang$^\ddagger$ \\\\
Doyen Sahoo$^\ddagger$\quad Junnan Li$^\ddagger$\quad Nan Jiang$^\dagger$\quad Tong Zhang$^\dagger$\quad Caiming Xiong$^\ddagger$\\\\
Hanze Dong$^*$$^\ddagger$\\\\ $^\ddagger$Salesforce AI Research\quad $^\dagger$University of Illinois Urbana-Champaign  }
\date{}
\usepackage{cancel}

\usepackage[skins]{tcolorbox} 
\tcbuselibrary{breakable} 
\newtcolorbox[auto counter, number within=section, list type=subsubsection, list inside=toc]{sectionbox}[2][]{
colback=white!98!gray, colframe=black, 
colbacktitle=white!90!gray, coltitle=black, 
fonttitle=\bfseries,
title={#2}, 
list entry={Comment \thetcbcounter\quad}
}
\usepackage[utf8]{inputenc}

\begin{document}

\maketitle

\def\thefootnote{*}\footnotetext{HD and WX contributed equally to this work. Corresponding to \texttt{hanze.dong@salesforce.com} and \texttt{wx13@illinois.edu}.}\def\thefootnote{\arabic{footnote}}

\begin{abstract}
 Reinforcement learning (RL) has become a prevailing approach for fine-tuning large language models (LLMs) on complex reasoning tasks. Among recent methods, GRPO stands out for its empirical success in training models such as DeepSeek-R1, yet the sources of its effectiveness remain poorly understood. In this work, we revisit GRPO from a reinforce-like algorithm perspective and analyze its core components. Surprisingly, we find that a simple rejection sampling baseline, RAFT, which trains only on positively rewarded samples, yields competitive performance than GRPO and PPO. 
Our ablation studies reveal that GRPO's main advantage arises from discarding prompts with entirely incorrect responses, rather than from its reward normalization.
Motivated by this insight, we propose \emph{Reinforce-Rej}, a minimal extension of policy gradient that filters both entirely incorrect and entirely correct samples. Reinforce-Rej improves KL efficiency and stability, serving as a lightweight yet effective alternative to more complex RL algorithms. We advocate RAFT as a robust and interpretable baseline, and suggest that future advances should focus on more principled designs for incorporating negative samples, rather than relying on them indiscriminately. Our findings provide guidance for future work in reward-based LLM post-training.
\end{abstract}

% \setlength{\parindent}{0pt}
% \setlength{\parskip}{8pt}
% \tableofcontents
% \newpage 

\section{Introduction} \label{sec:intro}
We investigate reinforcement learning (RL) algorithms in the context of fine-tuning large language models (LLMs) with verifiable rewards. Our focus is on mathematical reasoning tasks, which have recently received significant attention following the release of models such as OpenAI's O1 Model \citep{jaech2024openai} and DeepSeek-R1 \citep{deepseekai2025deepseekr1incentivizingreasoningcapability}. The dominant approach in LLM post-training has been Proximal Policy Optimization (PPO) \citep{schulman2017proximal, bai2022training, ouyang2022training}. However, PPO requires an additional critic network beyond the vanilla Reinforce algorithm \citep{williams1991function}, introducing both computational overhead and algorithmic complexity. Meanwhile, the deterministic transition nature of LLM also simplifies the problem with a relatively lower variance, many of PPO's sophisticated components may be unnecessary in this setting. This observation has inspired growing interest in designing simpler yet effective RL algorithms for post-training LLMs.

Several recent works revisit Reinforce-style approaches, including ReMax \citep{li2023remax}, RLOO \citep{ahmadian2024back, kool2019buy}, GRPO \citep{shao2024deepseekmath}, and Reinforce++ \citep{hu2025reinforce++}. In parallel, other methods explore different directions beyond policy gradients. Reward-ranked fine-tuning (RAFT) \citep{anthony2017thinking, dong2023raft} iteratively generates $n$ responses per prompt, filter out those with incorrect answers, and fine-tune the LLM on the remaining accepted samples. Direct preference-based methods, such as SFT-based contrastive learning (Slic-HF) \citep{zhao2023slic} and DPO \citep{rafailov2023direct}, optimize contrastive objectives based on a pairwise comparison dataset.

Among these, GRPO stands out as one of the most widely used algorithms for enhancing LLMs on math reasoning tasks due to its success in training DeepSeek-R1 \citep{deepseekai2025deepseekr1incentivizingreasoningcapability}. However, its algorithmic details remain largely undocumented, and it is unclear whether its adoption stems from inherent advantages or, rather, from continuity with methods used in their previous studies. To the best of our knowledge, a comprehensive justification of the algorithmic advantage of GRPO is still missing so far. In contrast, RAFT has established itself as one of the simplest and most interpretable baselines, consistently showing good empirical performance in prior literature despite its minimalistic design.

In this project, we revisit 
(1) RAFT, also know as rejection sampling in LLM literature, which is arguably the most basic RL algorithm for LLM post-training;
(2) Vanilla Reinforce, a classical policy gradient algorithm, serves as a simplified version of PPO by eliminating the critic model, and 
(3) GRPO, a Reinforce algorithm variant, samples $n$ responses per prompt and computes relative advantages by normalizing the sample reward using mean and standard deviation within each prompt.

A key difference between GRPO (Reinforce) and RAFT lies in how they handle negative samples: GRPO mixes both accepted and rejected examples during training, whereas RAFT relies only on positive samples. While it is commonly believed that RL methods leveraging negative signals significantly outperform SFT-like algorithms that only use positive samples, in our preliminary experiments, we observe that the performance gap is surprisingly small, and RAFT-like approach even exhibits faster convergence in the early training stage (e.g., the first 100-200 iterations). Our further analysis reveals that certain types of negative signals, such as prompts with entirely incorrect responses, can even can significantly hurt model performance. Meanwhile, other techniques like reward normalization appear to have minimal impact.

To better understand these dynamics, we conducted ablation studies isolating individual design choices using both Qwen \citep{qwen2} and LLaMA \citep{grattafiori2024llama} models across several Reinforce variants. Our results highlight the following key findings:
\begin{enumerate}
    \item We revisit RAFT, a simple rejection sampling baseline that uses only positive samples, and find that its performance is competitive with the state-of-the-art RL method GRPO with surprisingly small gap and faster convergence rate in the early training stages. A deeper analysis reveals that RAFT, which trains solely on positive samples, leads to a rapid reduction in policy entropy, limiting exploration and eventually being surpassed by GRPO.
    \item Through a set of controlled experiments across different Reinforce variants, we find that for on-policy methods, training on prompts where all sampled responses are incorrect can significantly harm performance. We further identify that the performance gain of GRPO over standard Reinforce largely stems from its implicit filtering of these harmful prompts. In contrast, reward normalization techniques by mean and standard deviation within a prompt have minimal impact.
    \item Motivated by our studies with both RAFT and Reinforce, we study a new Reinforce variant, \textit{Reinforce-Rej}, which selectively filters out prompts with either all correct or all incorrect responses. This method enjoys comparable final performance to GRPO, and demonstrates superior KL efficiency.
\end{enumerate}

These insights highlight the importance of sample selection over algorithmic design in reward-based LLM post-training. The codes of this project will be publicly available with detailed training scripts. The codes of this project are available at \url{https://github.com/RLHFlow/Minimal-RL}.

\section{Related Works}
Most of the related works on RL algorithm design for LLMs have been discussed in the introduction. Here, we review a few that are most relevant to our project.

\paragraph{Data filtering in LLM Post-Training.} Several recent works in RLHF and preference optimization explore data filtering strategies to improve training quality. For example, \citet{yuan2024self, dong2024rlhf, xiong2024building, shen2024policy} discard the candidates except for the top and bottom-ranked responses to reduce noise in pairwise comparisons during RLHF learning. \citet{yu2025rip} further incorporates reward and length information of rejected responses into the filtering process. For reasoning tasks, it is also common to remove prompts that are too easy or too hard \citep{yang2024qwen2,zhao2024automatic}, though this is typically done once before training. In contrast, our proposed Reinforce-Rej performs filtering online throughout training. Furthermore, our study reveals a connection between the strong empirical performance of GRPO and implicit data filtering. Reinforce-Rej can be seen as a natural extension of these insights from our ablation studies.

\paragraph{LLM for Mathematical Reasoning.} LLMs designed for (mathematical) reasoning have received significant attention, especially following the release of GPT-o1 by OpenAI \citep{jaech2024openai} and DeepSeek-R1 by DeepSeek \citep{deepseekai2025deepseekr1incentivizingreasoningcapability}. Earlier efforts primarily focused on building synthetic datasets and applying supervised fine-tuning \citep{gou2023tora, yue2023mammoth, yu2023metamath, toshniwal2024openmathinstruct}. In contrast, these new models (o1 and R1) adopt RL with verifier-based rewards as their main training approach. A key difference is that models like GPT-o1 and DeepSeek-R1 use more complex reasoning strategies—such as backward search and self-correction—and tend to generate longer outputs at inference time for better performance. Their success has inspired a surge of open-source efforts to replicate or adapt these training strategies to other domains \citep{jin2025search, xiong2025self,wang2024offline}. Notably, GRPO has become the default RL method in many of these projects, often without justification. However, whether GRPO is truly better than Reinforce, and (if the answer to the first question is yes) what contributes to its performance gains, remains largely under-explored.

\section{Method}

\paragraph{Notation.} 
Given a prompt, an LLM is denoted as a policy that can map the prompt to a distribution over response $a$: $\pi(a|x)$. We also denote $r(x,a) \in \{-1, 1\}$ as a binary reward function that assigns scalar feedback to a prompt-response pair, which can be implemented by the verifier\footnote{\url{https://github.com/huggingface/Math-Verify}}. We denote the dataset of collected prompt-response pairs as $\mathcal{D}$. For each prompt $x$, we can generate $n$ candidate responses $a_1, \cdots, a_n$, and their corresponding rewards are $r_1, \cdots, r_n$.

Let $a_t$ be the $t$-th token in response $a = (a_1, \cdots, a_{|a|})$, and let $s_t(\theta) = \frac{\pi_{\theta}(a_t|x, a_{1:t-1})}{\pi_{\theta_\text{old}}(a_t|x, a_{1:t-1})}$ denote the importance sampling ratio for token $t$. We also define the baseline of rewards as $\mathrm{mean}(r_1, \cdots, r_n)$ and its standard deviation as $\mathrm{std}(r_1, \cdots, r_n)$.  We now review several representative algorithms used for the LLM post training.

%We assume that we have access to a binary reward function $r(x,a) \in \{0, 1\}$, which can assign a scalar reward to any prompt-response pair $(x,a)$. We now review the representative algorithms used for the LLM post training.

\paragraph{RAFT.} The RAFT algorithm is also referred to as the rejection sampling fine-tuning \citep{touvron2023llama, yuan2023scaling} in the literature. We follow the formalization in \citet{dong2023raft}, which consists of the following three steps:
\begin{itemize}
    \item \textbf{Data Collection.} For a batch of prompts $\{x_1, \cdots, x_M\}$, we sample $n$ responses per prompt from a reference model (e.g., the current model) to obtain candidate responses $\{a_{i,1}, \cdots, a_{i,n}\}$ for each $x_i$.
    \item \textbf{Data ranking (Rejection Sampling).} For each prompt $x_i$, we compute the reward values of each response $\{r_{i,1}, \cdots, r_{i,n}\}$ using the binary reward function $r(x,a)$, and retain only the responses with the highest reward (typically those with $r=1$). The resulting set of positive samples is aggregated into a dataset $\mathcal{D}$.
    \item \textbf{Model Fine-Tuning.} The current policy $\pi$ is then fine-tuned to maximize the log-likelihood over the selected dataset:
    \begin{equation} \label{eqn:raft_loss}
\mathcal{L}^{\text{RAFT}}(\theta) = \sum_{(x,a) \in \mathcal{D}} \log \pi_\theta(a|x).
    \end{equation}
    %\item \textbf{Model fine-tuning.} We then fine-tune the current language model $\pi$ by maximizing the log likelihood
    % \begin{equation} \label{eqn:raft_loss}
    %  \sum_{(x,a) \in \cD} \log \pi(a|x).
    % \end{equation}
\end{itemize}
A closely related algorithm is STaR \citep{zelikman2022star}, which also trains on self-generated CoT responses. In comparison, STaR retrains from the original pre-trained model in each iteration instead of the current model. Meanwhile, STaR uses greedy decoding and generate only one response, as compared to the rejection sampling used in RAFT. Lastly, STaR also proposes to provide the answer in the prompt to generate CoT responses for difficult problems.

\paragraph{Policy Gradient and Reinforce.} We illustrate the idea by taking the action as a whole for simplicity and extend to the autoregressive model later. The policy gradient algorithm is designed to solve the following learning objective:
\begin{equation}
    J(\theta) = J(\pi_\theta) = \E_{x \sim d_0} \big[\E_{a \sim \pi_\theta(\cdot|x)} r(x, a) \big],
\end{equation}
where $\theta$ is the parameter of the neural network. We can use policy ascent to update the policy network:
$$
\theta' \leftarrow \theta + \beta \cdot \nabla_\theta J(\theta), 
$$
where $\nabla_\theta J(\theta)$ is referred to as the \textbf{policy gradient} in the literature. The policy gradient is given by:
$$
\frac{\partial J(\theta)}{\partial \theta} = \E_{x\sim d_0} \Big[\E_{a \sim \pi_\theta(\cdot|x)} \big[\frac{\partial \log \pi_\theta(a|x)}{\partial \theta} \cdot r(x,a)\big] \Big].
$$
In practice, similar to the pipeline of RAFT, we usually use $\pi_{\theta_{\text{old}}}$ to collect the trajectories into the replay buffer $\cD$ and use these samples to compute a stochastic policy gradient to update $\pi_{\theta_{\text{old}}}$. However, for a strict on-policy training, we have to collect new data after a single step of gradient ascent. To accelerate training, we usually perform multiple steps in a mini-batch manner, and adopt the importance sampling technique to correct the distribution. Specifically, we can re-write the objective function as:
\begin{equation} 
    J(\theta) = J(\pi_\theta) = \E_{x \sim d_0} \Big[\E_{a \sim \pi_{\theta_{\text{old}}}(\cdot|x)} \big[\frac{\pi_{\theta}(a|x)}{\pi_{\theta_{\text{old}}}(a|x)}r(x, a)\big] \Big].
\end{equation}
Then, with a batch of trajectories $\{x, a, r\}$ collected by $\pi_{\theta_{\text{old}}}$, we can update multiple steps using the above importance sampling trick. However, the importance sampling can lead to high variance if the distribution of $\pi_{\theta}$ and $\pi_{\theta_{\text{old}}}$ are too far away. To stabilize the training, we can also leverage the clipping techniques from the PPO. Finally, the loss function is:
\begin{equation} \label{eqn:sentence_level}
    \mathcal{L}^{\text{Reinforce}}(\theta) = \frac{1}{|\cD|} \sum_{x,a \in \mathcal{D}} \Big[ \min\Big(\frac{\pi_{\theta}(a|x)}{\pi_{\theta_{\text{old}}}(a|x)}r(x, a), \mathrm{clip}( \frac{\pi_{\theta}(a|x)}{\pi_{\theta_{\text{old}}}(a|x)}, 1-\epsilon,1+\epsilon)\cdot r(x,a)\Big)\Big].
\end{equation}
Since LLM is autoregressive, we typically regard each token as an action. Therefore, we can extend the loss to the token-level counterpart:
\begin{equation} \label{eqn:token_level}
        \mathcal{L}^{\text{Reinforce}}(\theta) = \frac{1}{|\cD|} \sum_{x,a \in \mathcal{D}} \frac{1}{|a|}\sum_{t=1}^{|a|} \Big[ \min\Big(s_t(\theta), \mathrm{clip}(s_t(\theta), 1-\epsilon,1+\epsilon)\cdot r (x,a)\Big)\Big],
\end{equation}
where $s_t(\theta) = \frac{\pi_{\theta}(a_t|x, a_{1:t-1})}{\pi_{\theta_\text{old}}(a_t|x, a_{1:t-1})}$ and $a_t$ is the $t$-th token of $a$.

%\paragraph{Reinforce+.}

\paragraph{GRPO.} GRPO adopts a loss function similar to \eqref{eqn:token_level}, but replaces $r(x,a)$ with an advantage function $A_t(x,a)$ for the $t$-th token of response $a$. Specifically, for each prompt $x$, GRPO will sample $n > 1$ responses and compute the following advantage for the $t$-th token of the i-th response:
$$
A_{t}(x,a_i) = \frac{r_{i} - \mathrm{mean} (r_1, \cdots r_n)}{\mathrm{std}(r_1,\cdots,r_n)}. 
$$
$\mathrm{mean} (r_1, \cdots r_n)$ is often referred to as the baseline in the RL literature, which serves to reduce the variance of the stochastic gradient.

\paragraph{(Iterative) DPO.} The DPO algorithm relies on pairwise comparison dataset $\{(x, a^+, a^-)\}$, where $a^+ \succ a^-$ are two responses to the prompt $x$. Then, DPO optimizes the following contrastive loss:
$$
\mathcal{L}^{\text{DPO}}(\theta) = - \log \sigma \Big(\beta \log \frac{\pi_\theta(a^+|x)}{\pi_{\mathrm{ref}}(a^+|x)} - \beta \log \frac{\pi_\theta(a^-|x)}{\pi_{\mathrm{ref}}(a^-|x)}\Big),
$$
where $\beta > 0$ and $\pi_{\mathrm{ref}}$ is usually set as the initial checkpoint. The original DPO algorithm trains on offline and off-policy data. In the subsequent studies \citep{liu2023statistical, xiong2023iterative, xu2023some, snorkelai@pair, dong2024rlhf}, it is shown that we can iteratively use the intermediate checkpoints to generate new responses, label the preference signals, and train on the self-generated on-policy data to largely improve the model performance.

\paragraph{RAFT++.} We notice that RAFT can also be viewed as a hybrid algorithm that can be off-policy when performing multiple steps on the replay buffer at each iteration. As a natural extension, we also apply the importance sampling and clipping techniques to the original RAFT, arriving at a similar loss function:
\begin{equation} \label{eqn:token_level_raftpp}
\begin{aligned}
            \mathcal{L}^{\text{RAFT++}}(\theta) &= \frac{1}{|\cD|} \sum_{x,a \in \mathcal{D}} \frac{1}{|a|}\sum_{t=1}^{|a|} \Big[ \min\Big(s_t(\theta), \mathrm{clip}(s_t(\theta), 1-\epsilon,1+\epsilon)\Big)  \mathcal{I}\big( r(x,a) = \argmax_{i} r(x, a_i) \big)\Big],
            \end{aligned}
\end{equation}
where the indicator ensures that we only train on the response with the highest reward (positive samples).

\section{Experiment Setup}

We focus on the mathematical reasoning task in this project. The implementations are mainly based on the verl \citep{sheng2024hybridflow} framework. 

\begin{table*}[t]
    \centering
    \begin{adjustbox}{max width=\textwidth}
    \begin{tabular}{cc|cccc}
   \toprule
        Model & Algorithm & \textbf{Math500} & \textbf{Minerva Math} & \textbf{Olympiad Bench} & \textbf{Average} \\
        \midrule
        \multirow{1}{*}{Qwen2.5-Math-7B-base} & Base & 41.3 & 11.0 & 18.6 & 23.6 \\
        & RAFT & 77.4 & 40.8 & 38.6 & 52.3 \\
       \rowcolor{cyan!20}  & RAFT++ & 80.2 & 44.9 & 43.3 & 56.1 \\
        & Iterative DPO & 76.0 & 31.2 & 39.3 & 48.8\\
        & Reinforce & 80.1 &	40.7	& 40.9		&		53.9 \\
       \rowcolor{cyan!20} & GRPO & 81.3 & 45.5 & 42.2 & 56.3 \\
                & PPO & 79.0 & 39.3 & 39.1 & 52.5\\
      \rowcolor{cyan!20}  & Reinforce-Rej & 81.9 &	44.2&	43.1 & 56.4 \\
        \midrule
        \multirow{1}{*}{LLaMA-3.2-3B-instruct} & Base & 26.3&	7.4&	5.5 & 13.1 \\
& RAFT & 46.1 &	17.6 &	13.9 & 25.9 \\
      \rowcolor{cyan!20}  & RAFT++ & 47.4	 &19.1&	16.3 & 27.6 \\
        & Reinforce & 45.9 & 13.7 & 13.0 & 24.2 \\
        \rowcolor{cyan!20}& GRPO & 49.2 &	19.3&	16.8 & 28.4 \\
                & PPO & 46.5	& 19	& 15.1 & 26.9 \\
         \rowcolor{cyan!20}   & Reinforce-Rej & 50.1 &	19.3	& 16.1 & 28.5\\
\bottomrule
    \end{tabular}
    \end{adjustbox}
    \caption{Performance of different algorithms across three benchmarks including Math500 \citep{hendrycks2021measuring}, Minerva Math \citep{lewkowycz2022solving}, and Olympiad Bench \citep{he2024olympiadbench}. We tune all the algorithms to their best performance by fully optimizing the hyper-parameters (including batch size, mini batch size, and actor learning rate).  See Appendix for the detailed parameter setup.  The reported accuracy is average@16 with a temperature 1.0 and a maximal generation length of 4096 tokens. }
    \label{tab:main res}
\end{table*}

\paragraph{Dataset and Models.} We train the models using the prompt set Numina-Math \citep{numina_math_7b}, which consists of approximately 860k math problems and labeled ground-truth answers. The sources of Numina-Math ranges from Chinese high school math exercises to US and international mathematics olympiad competition problems. We conduct experiments with both Qwen2.5-Math-7B-base, and LLaMA-3.2-3B-instruct for generality. We use the default chat template of these models and use CoT prompting: ``Let's think step by step and output the final answer within \textbackslash boxed\{\}''.

\paragraph{Hyper-parameters.} We follow most of the hyper-parameter setups recommended in the verl framework for the Reinforce, GRPO, and PPO training. The hyper-parameters for RAFT and RAFT++ are also the same with the GRPO script. Specifically, we use the AdamW optimizer with a learning rate of $1\times 10^{-6}$. We sample $1024$ prompts per iteration, and generate $n=4$ responses per prompt for RAFT and GRPO. The training mini-batch size is set to be $512$. The models are allowed to generate 4096 tokens at most during training. More detailed scripts are available in the GitHub repository. For the baseline of iterative DPO, we use the codebase developed in \citet{zhang2025online}.

\paragraph{Evaluation.} We evaluate the models' reasoning ability by Math500 \citep{hendrycks2021measuring}, Minerva Math \citep{lewkowycz2022solving}, Olympiad Bench \citep{he2024olympiadbench}. We do not include the popular AIME2024 benchmark since it only consists of 30 problems. In our preliminary experiments, we observe that the trend on this benchmark is very noisy for all the considered algorithms. We mainly use average@16 to evaluate our models, where we generate $16$ responses per prompt with temperature 1.0, and use the average accuracy as the metric. The models are allowed to generate 4096 token at most.

The codes are available at \url{https://github.com/RLHFlow/Minimal-RL}.

\section{Main Results}
 %\rowcolor{cyan!20} 

\paragraph{RAFT and RAFT++ approach deep RL methods with surprisingly small performance gap.} We summarize the test accuracy of models trained using various algorithms in Table~\ref{tab:main res}. Our first observation is that RAFT (and its variant RAFT++), which is arguably the simplest algorithm, achieves competitive performance compared to more complex methods such as iterative DPO and deep RL-based approaches. Specifically, with Qwen2.5-Math-7B-base, vanilla RAFT reaches an average accuracy of 52.3\%, outperforming iterative DPO (48.8\%) and approaching PPO (52.5\%). With the additional importance sampling and clipping techniques, RAFT++ further improves over vanilla RAFT, achieving 56.1\% average accuracy. This result is remarkably close to the state-of-the-art deep RL method GRPO, which reaches an average accuracy of 56.3\% in its best model. A similar trend is observed on the LLaMA-3.2-3B-instruct model, demonstrating the robustness of RAFT and RAFT++ across different models. These results are somewhat counter-intuitive, as RL methods are often believed to be more powerful due to their ability to utilize negative feedback. Interestingly, in the LLaMA-based setting, Reinforce performs substantially worse than RAFT++, with an average accuracy of 24.2\% compared to the 27.6\% of RAFT++. 

One possible explanation is that defining negative samples solely based on final answer correctness may be too coarse, potentially limiting the benefits of negative signals. Moreover, Reinforce with binary reward ($\pm 1$) can be viewed as fine-tuning on the positive samples and unlearning on the negative samples. When the negative signals are not sufficiently fine-grained, unlearning on the negative samples is more unstable than fine-tuning on the positive samples. We will also include more ablation studies to investigate the role of the negative samples in current practice of RL training.

\begin{figure}[H]
    \centering
    \includegraphics[width=0.48\textwidth]{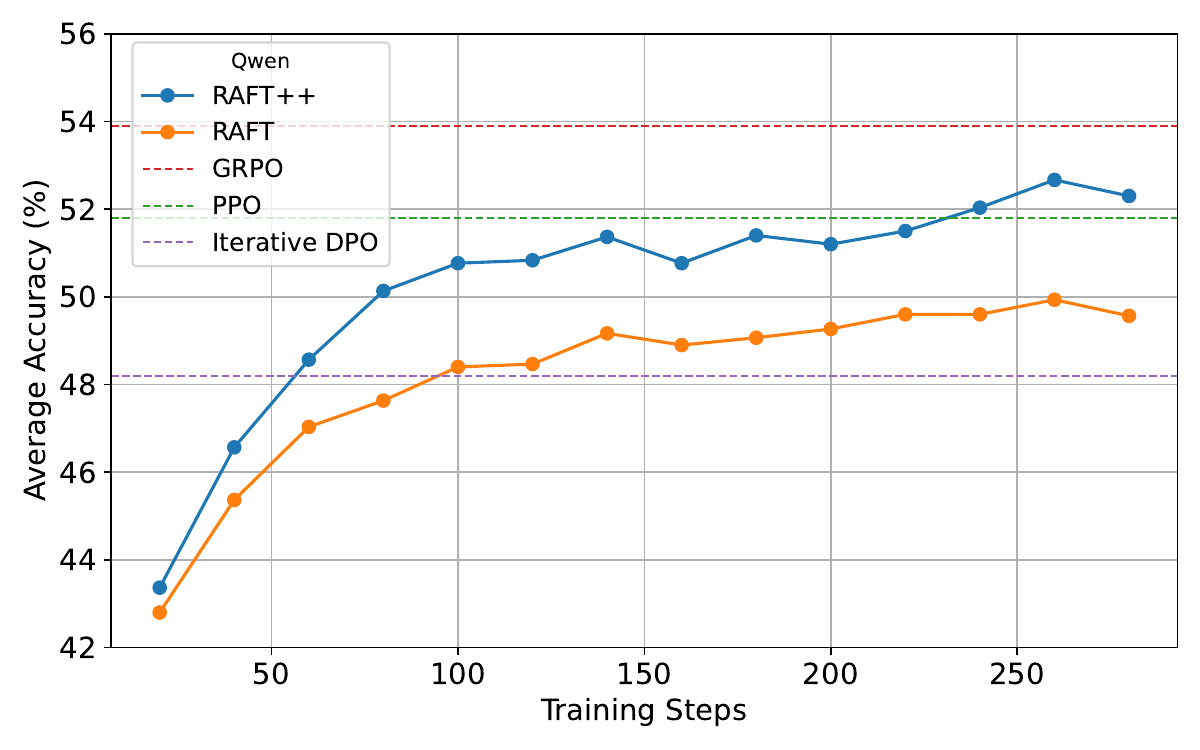}  
    \includegraphics[width=0.48\textwidth]{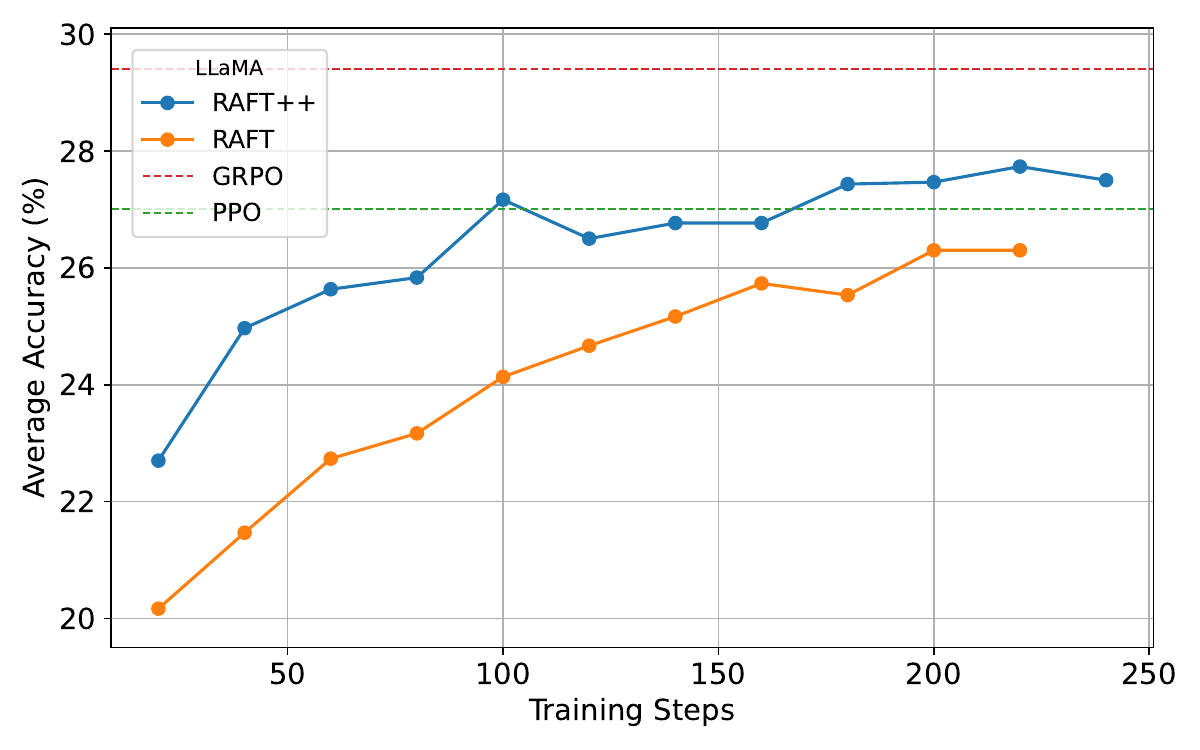} 
    \caption{The learning dynamics of RAFT and RAFT++, initialized from Qwen2.5-Math-7B-base (left) and LLaMA-3.2-3B-instruct (right). The y-axis is the average@16 accuracy, that is further averaged on MATH500, Minerva Math, and Olympiad Bench. We also plot the best model of GRPO, PPO, and Iterative DPO for reference.}
        \label{fig:ppo} 
\end{figure}

\paragraph{Distribution correction and clipping improve vanilla RAFT.}
Table~\ref{tab:main res} also shows that applying importance sampling to correct for distribution shift in the replay buffer improves the final test accuracy of RAFT, leading to a stronger variant we refer to as RAFT++. We further illustrate the learning dynamics of RAFT and RAFT++ in Figure~\ref{fig:ppo}. Both methods are capable of steadily enhancing the model's reasoning ability through online updates, with RAFT++ achieving faster convergence and higher final accuracy than vanilla RAFT. 

\begin{figure}[H]
    \centering
    \includegraphics[width=0.46\textwidth]{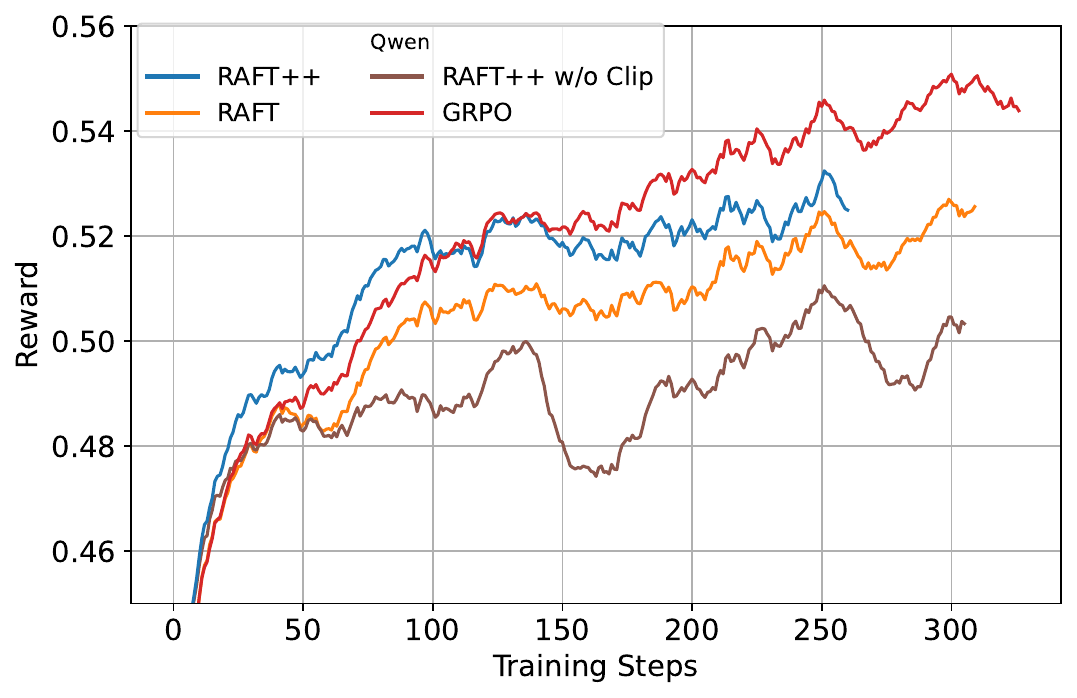}  
    \includegraphics[width=0.48\textwidth]{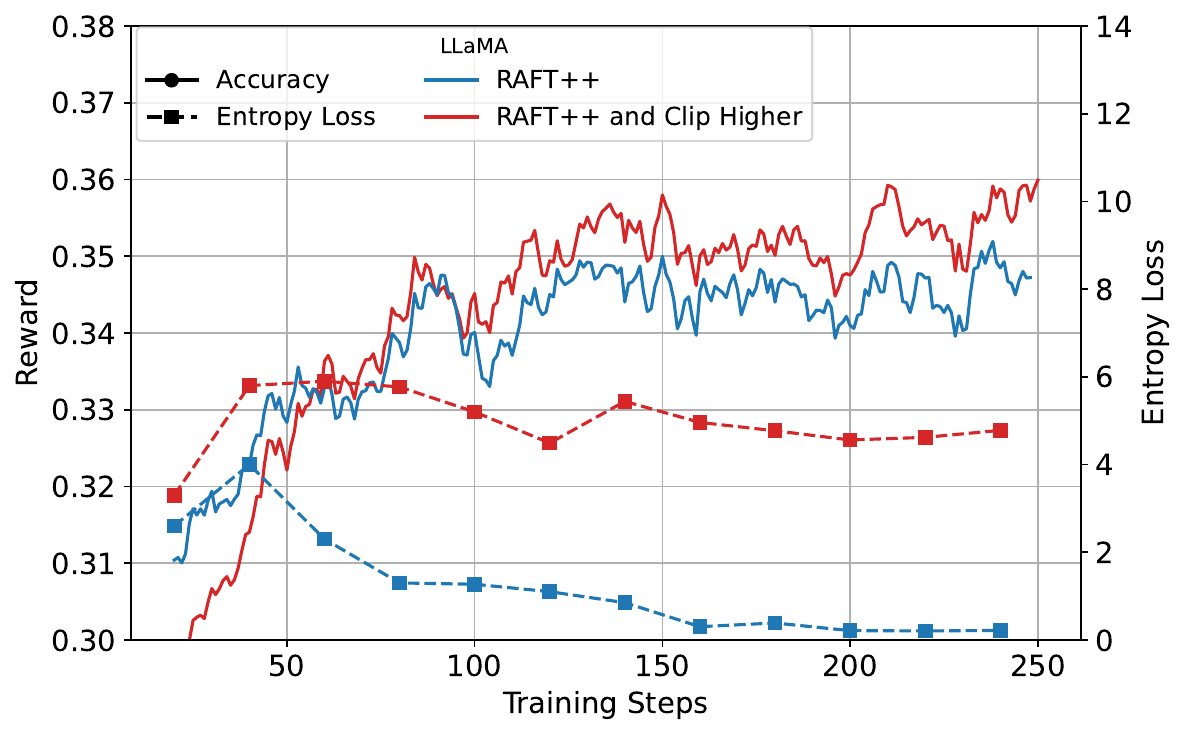}
    \caption{Left: the training reward curves of RAFT, RAFT++, RAFT++ without clipping (i.e., RAFT and importance sampling), and GRPO, initialized from Qwen2.5-Math-7B-base. Right: the training reward curves of RAFT++ and RAFT++ enhanced by clip higher trick, initialized from LLaMA-3.2-3B-instruct. We transform the original reward using $(1 + r)/2$ so that the resulting value corresponds to the accuracy on the training data. We also apply a moving average with a window size of $20$ to smooth the curves.
    }
        \label{fig:training_reward_qwen} 
\end{figure}

As part of our ablation study, we also evaluate an intermediate variant that applies importance sampling without clipping. As shown in Figure~\ref{fig:training_reward_qwen}, this variant underperforms vanilla RAFT. This observation contradicts the findings of \citet{ahmadian2024back}, which suggest that clipping rarely occurs and is therefore unnecessary. We hypothesize that although clipping may be infrequent, it happens when $\frac{\pi_{\theta}}{\pi_{\theta_{\text{old}}}}$ deviates far from $1$. In such cases, unbounded updates can severely violate the on-policy assumption underlying policy gradient methods, leading to instability and degraded performance.

\paragraph{RAFT++ achieves faster early-stage convergence but is surpassed by GRPO in later training.} From Figure~\ref{fig:training_reward_qwen}, we also observe that RAFT++ exhibits faster early-stage learning compared to GRPO. Moreover, we also observe a clear turning point in the training dynamics around iteration 100, where its growth rate slows noticeably after this point. Eventually, RAFT++ is surpassed by GRPO in the later stage of training in terms of final model test accuracy. We will also conduct ablation experiments to investigate the cause of this slowdown in RAFT++ and the role of the missing negative samples in this process.

\subsection{Ablation Study}
In this subsection, we aim to understand the underlying reasons behind the key findings presented earlier. To this end, we conduct a series of ablation studies designed to answer the following questions: 
\begin{enumerate}
    \item From RAFT++ to Reinforce (including GRPO): Why is RAFT++ faster in the early stage but ultimately outperformed later in training? What role do negative samples play? 
    \item From Vanilla Reinforce to GRPO: What is the key factor behind GRPO's superior performance? 
\end{enumerate} 

\begin{figure}[H]
    \centering
    \includegraphics[width=0.48\textwidth]{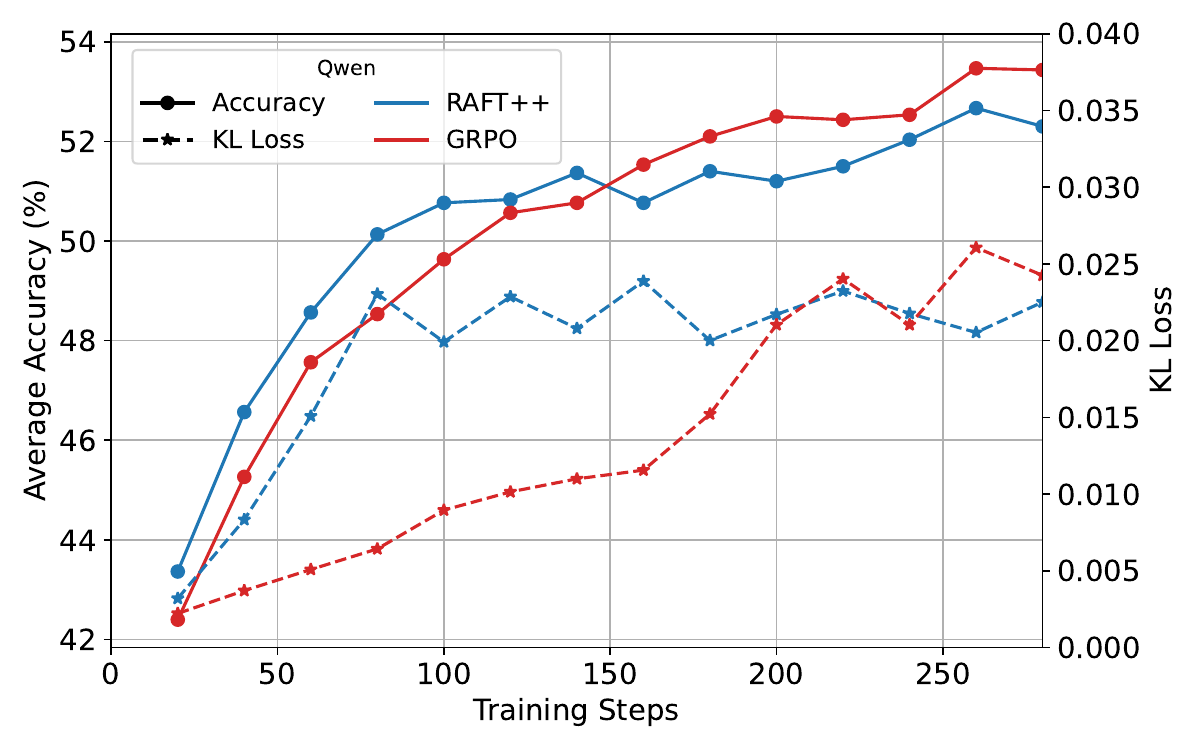}  
    \includegraphics[width=0.48\textwidth]{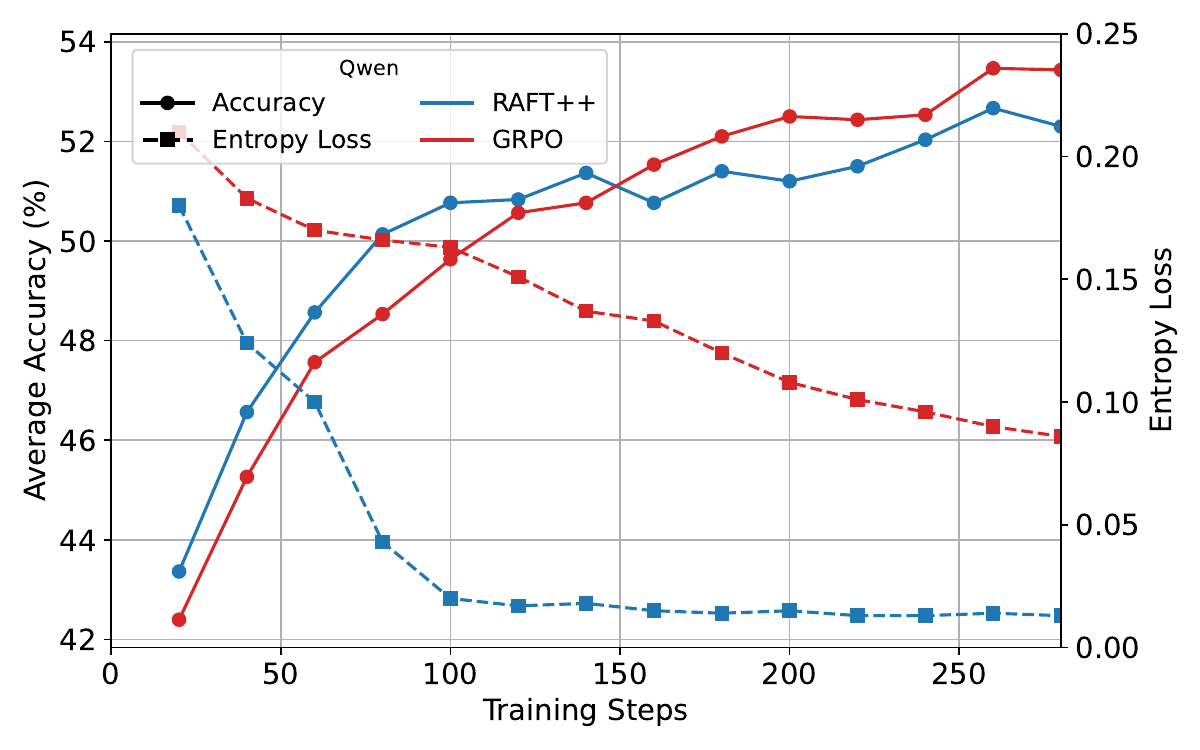} 
        \includegraphics[width=0.48\textwidth]{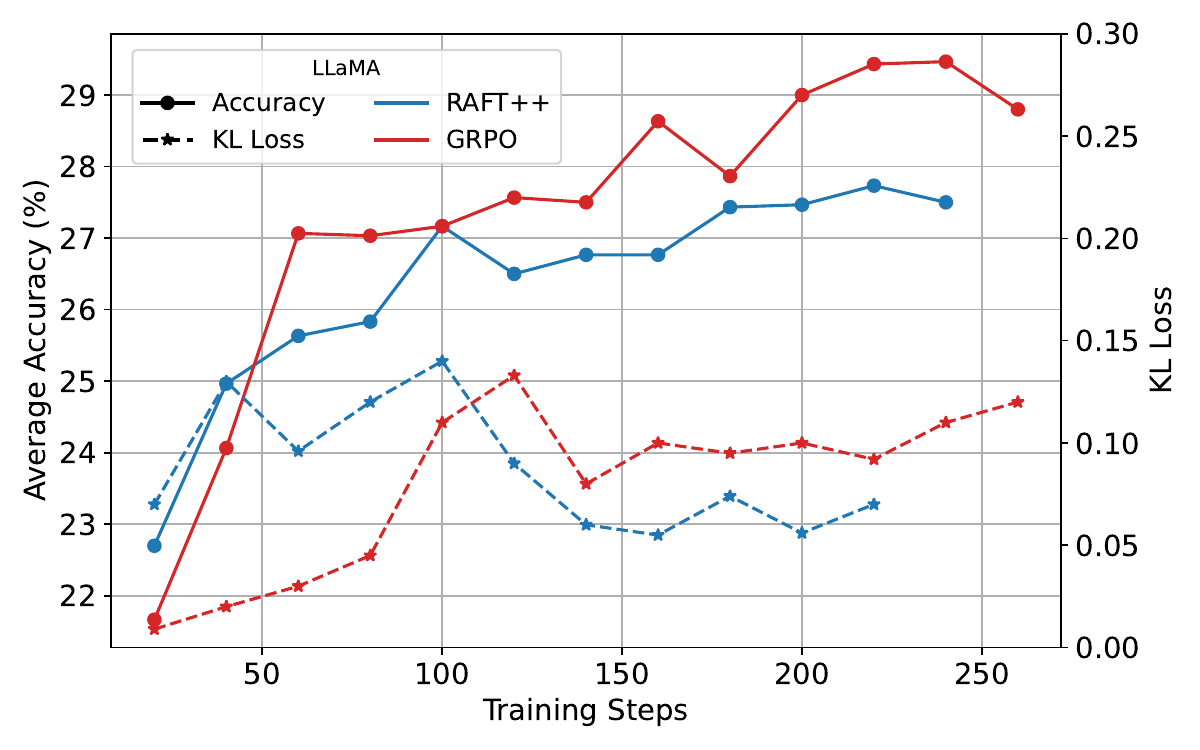}  
    \includegraphics[width=0.48\textwidth]{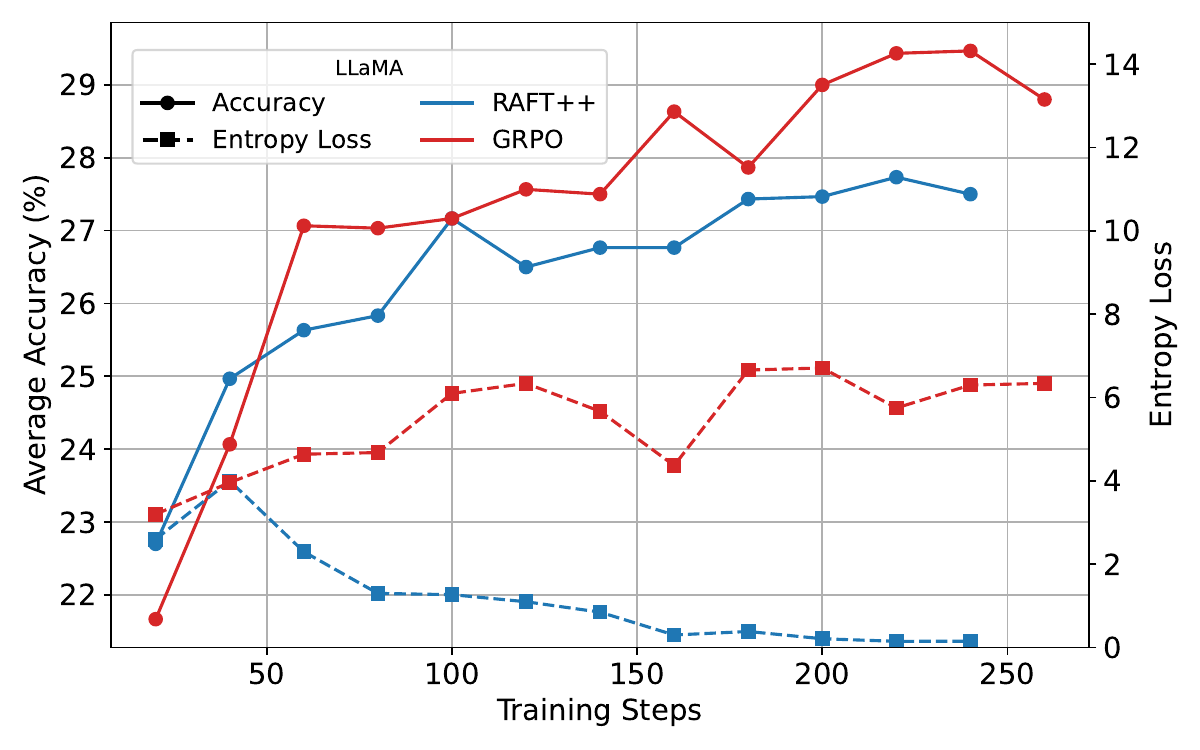} 
    \caption{The learning dynamics of RAFT++ and GRPO, initialized from Qwen2.5-Math-7B-base (first row) and LLaMA-3.2-3B-instruct (second row). We also plot the KL loss in the left column and the policy entropy loss in the right column. }
        \label{fig:ablation_entropy} 
\end{figure}

\paragraph{Learning from only positive samples leads to faster convergence and entropy collapse.} We begin by examining the policy entropy and KL divergence from the initial policy for RAFT++ and GRPO, as shown in Figure~\ref{fig:ablation_entropy}. A key observation is that RAFT++, which trains exclusively on positive samples, exhibits a much more rapid decline in policy entropy compared to GRPO. This trend is consistent across both Qwen and LLaMA models. Once the entropy stabilizes at a low level, the performance improvement of RAFT++ slows noticeably. We attribute this to reduced exploration with the low-entropy policies, since they are less likely to generate diverse reasoning paths. In parallel, the KL divergence from the initial policy increases more rapidly in RAFT++ during early training, reflecting its initial advantage in test accuracy. However, due to the lack of continued exploration, RAFT++ quickly plateaus, while GRPO continues to improve and ultimately surpasses it. 

These findings suggest that negative samples play a crucial role in maintaining exploration and preventing distributional collapse. This exploration benefit is likely a contributing factor to the performance gap between RAFT++ and RL-based methods such as Reinforce and GRPO. To further investigate the relationship between policy entropy and reward learning, we incorporate the ``clip higher'' technique from \citet{yu2025dapo}, which uses an asymmetric clipping range with $\epsilon_1 = 0.2$ for the lower bound and a larger $\epsilon_2 = 0.28$ for the upper bound. We apply this variant to the LLaMA-3.2-3B-instruct model and visualize both the training reward curves and policy entropy curves in the right figure of Figure~\ref{fig:training_reward_qwen}. Consistent with the findings in \citet{yu2025dapo}, using a larger $\epsilon_2$ helps stabilize the policy entropy over the online training. As a result, this enhanced RAFT++ variant outperforms the original RAFT++ during the later stages of training.

\paragraph{From Reinforce to GRPO: what is the key role to the success of GRPO?}

The primary differences between GRPO and RAFT lie in two aspects: the use of negative samples and the application of reward normalization. To isolate the contributions of each component and better understand their respective effects, we designed a set of controlled experiments to systematically evaluate their impacts. Specifically, we consider the following algorithms:
\begin{enumerate}
    \item Reinforce: the vanilla one introduced in \eqref{eqn:token_level};
    \item Reinforce + Mean Zero: we subtract the mean reward within each prompt;
    \item Reinforce + Remove all correct: we filter out prompts whose responses are entirely correct;
    \item Reinforce + Remove all wrong: we filter out prompts whose responses are entirely wrong;
    \item Reinforce + Remove both: remove both fully correct and fully incorrect prompts;
    \item Reinforce + Remove both + Normalized Std: in addition to removing both fully correct and fully incorrect prompts, we further divide the reward by its standard deviation within each prompt for normalization.
\end{enumerate}

\begin{figure}[htp]
    \centering
    \includegraphics[width=0.32\textwidth]{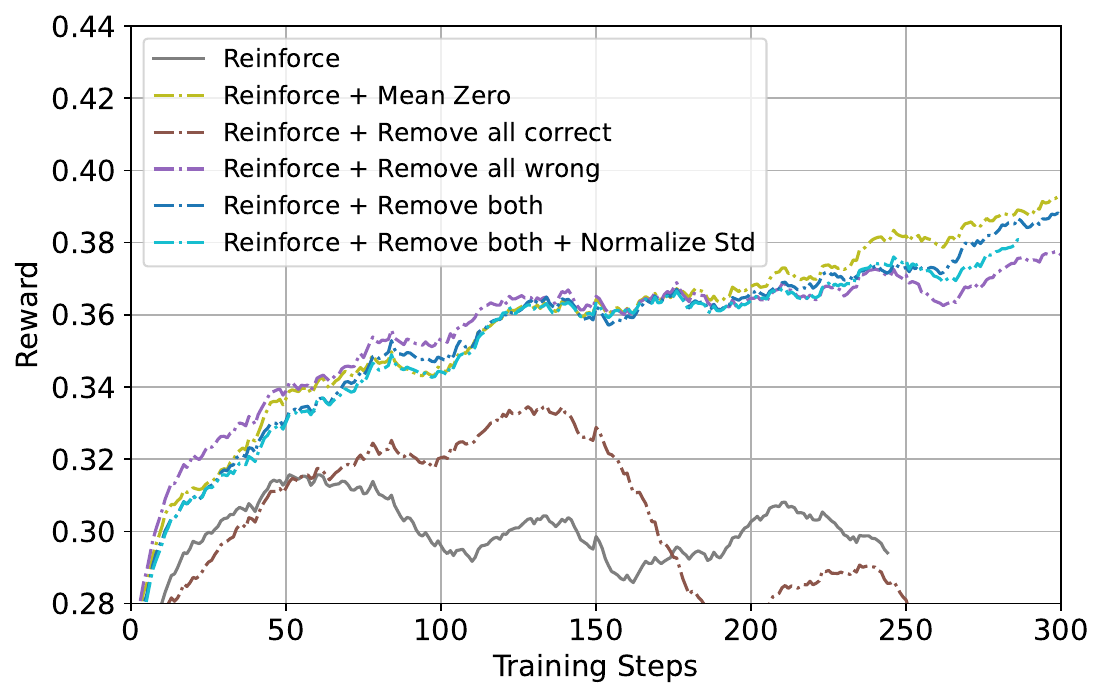}  
    \includegraphics[width=0.32\textwidth]{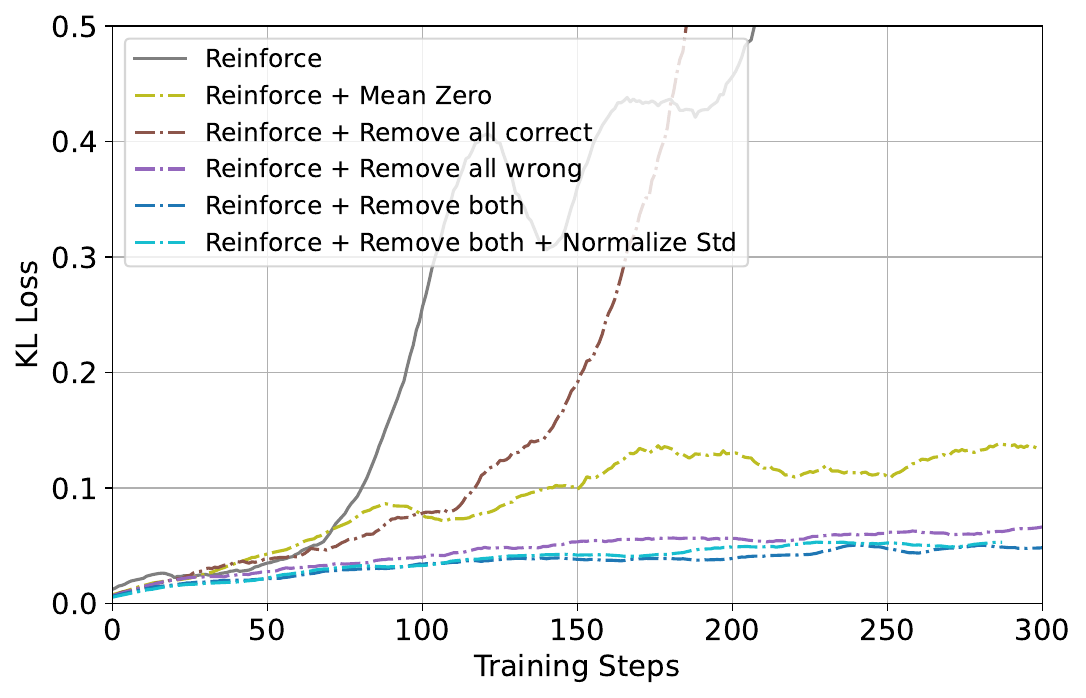} 
    \includegraphics[width=0.32\textwidth]{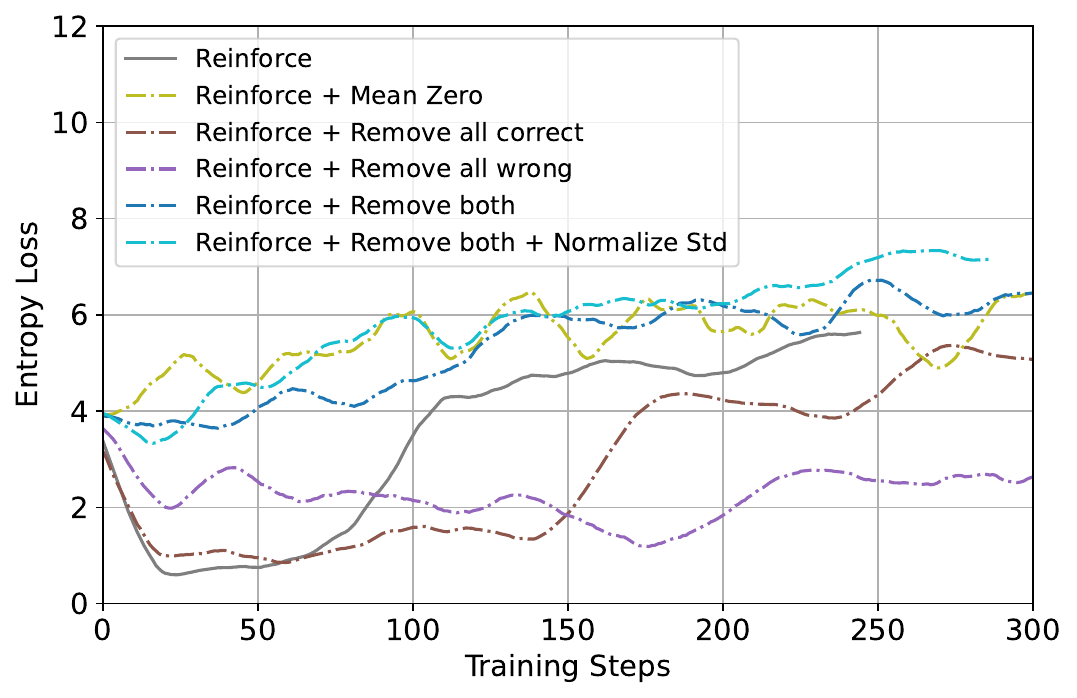}
    \includegraphics[width=0.32\textwidth]{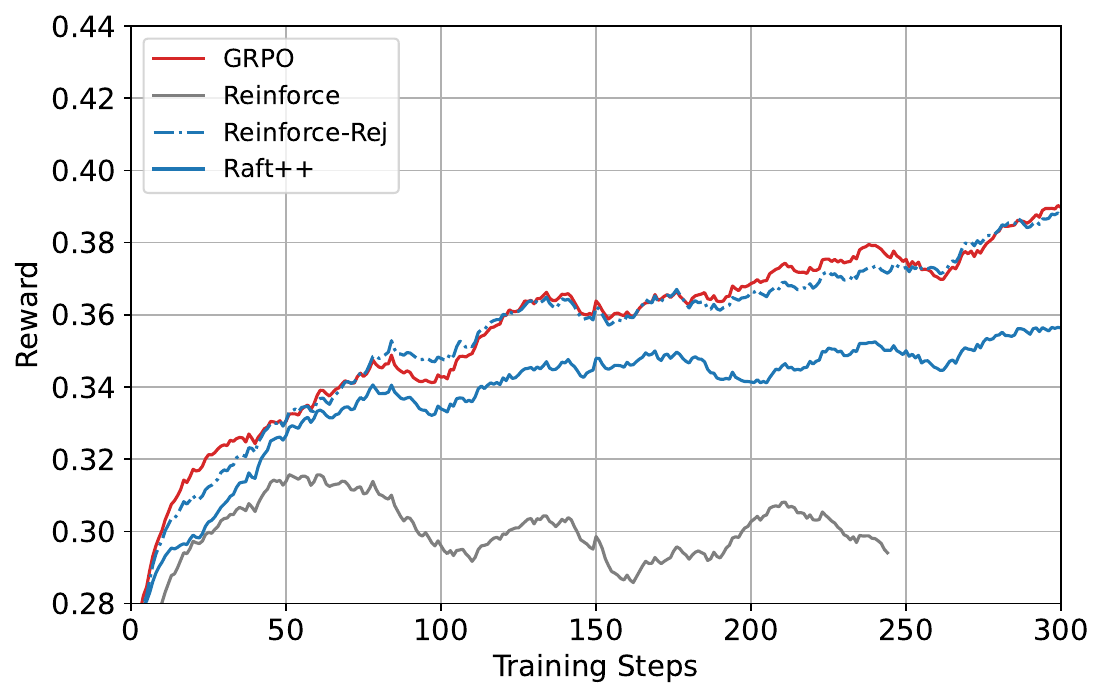}  
    \includegraphics[width=0.32\textwidth]{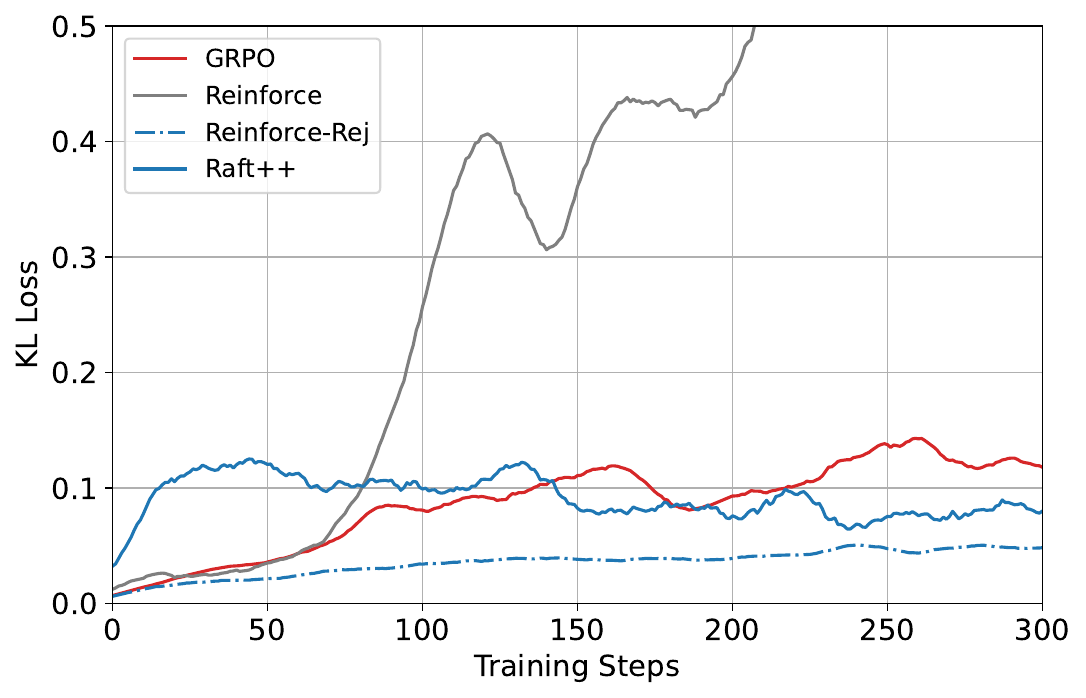} 
    \includegraphics[width=0.32\textwidth]{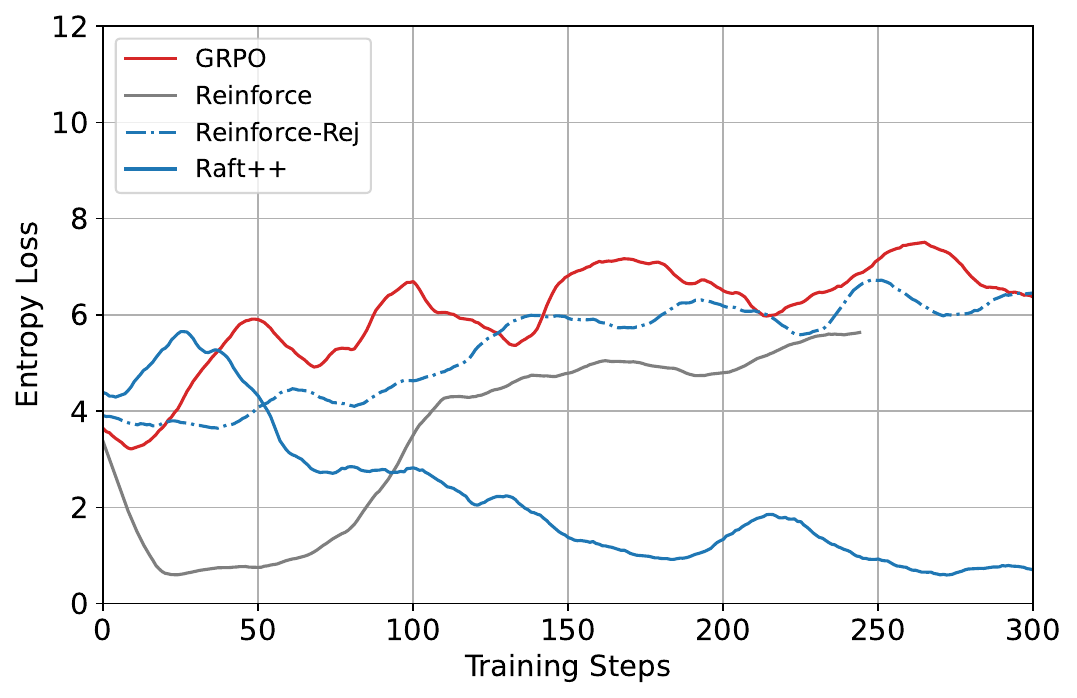}
    \caption{Ablation study on the components of GRPO and Reinforce-type algorithms with LLaMA-3.2-3B-instruct. We compare GRPO with other Reinforce-based variants to isolate the effects of removing incorrect samples, correct samples, and applying normalization.  Removing incorrect samples (“Remove all wrong”) provides the largest gain in reward, highlighting their harmful impact. In contrast, the reward of removing correct samples is still not satisfactory. Mean-zero normalization increases KL loss and destabilizes training. Normalizing by standard deviation shows minimal additional benefit. The variant “Reinforce + Remove both” achieves a good balance between reward, KL stability, and entropy regularization. We transform the original reward using $(1 + r)/2$ so that the resulting value corresponds to the accuracy on the training data. We also apply a moving average with a window size of $20$ to smooth the curves.}
        \label{fig:ablation_grpo}
\end{figure}

As shown in Figure~\ref{fig:ablation_grpo}, the variant ``Reinforce + Remove all wrong'' achieves the significant performance improvement than vanilla Reinforce in terms of reward, clearly indicating that incorrect samples are particularly harmful in the Reinforce training process. This is likely due to their high variance and misleading gradients, which can dominate updates and misguide learning. In contrast, removing only correct samples (“Reinforce + Remove all correct”) does not help much.
Meanwhile, removing both all correct and wrong samples result in more well-behaved entropy loss and slightly better reward, suggesting that it can help maintain exploration.

We also observe that normalization alone, such as in the “Reinforce + Mean Zero” variant, leads to increased KL divergence and does not improve reward, indicating potential instability. Moreover, applying standard deviation normalization (“Reinforce + Remove both + Normalize Std”) yields little additional gain over simply removing bad samples, suggesting that variance normalization is not a key contributor to performance.

Taken together, these results highlight that the core strength of GRPO lies in rejecting low-quality (especially incorrect) samples, rather than normalization per se. We refer to the variant that removes both correct and incorrect samples--``Reinforce + Remove both''--as Reinforce-Rej, which serves as a simplified yet competitive baseline for reward-based policy optimization in LLMs.

\section{Conclusion}

We revisited the design space of reinforcement learning algorithms for LLM post-training through the lens of rejection sampling. Our study shows that RAFT—a simple rejection-based method relying solely on positively rewarded samples—serves as a surprisingly strong baseline, outperforming or matching more sophisticated approaches such as PPO and iterative DPO. We further improved RAFT by incorporating importance sampling and clipping, resulting in RAFT++, which achieves near state-of-the-art performance while maintaining a simple and stable training pipeline.

Through extensive ablations, we identified that GRPO's primary benefit comes not from its reward normalization, but from discarding prompts with entirely correct and incorrect responses. Building on this insight, we proposed Reinforce-Rej, a minimal policy gradient variant that filters both entirely incorrect and entirely correct samples. Reinforce-Rej improves KL efficiency and entropy stability, highlighting the role of exploration in reward-based fine-tuning.

Our findings suggest that the utility of negative samples in RL-based LLM training is more nuanced than previously assumed. Rather than relying on raw negative feedback, future methods should consider more selective and principled mechanisms for incorporating sample quality. We advocate RAFT and Reinforce-Rej as lightweight, interpretable, and effective baselines for future work on reward-driven LLM post-training.

\bibliography{myrefs}
\bibliographystyle{apalike}

% \newpage
% \appendix

\end{document}